\documentclass[letterpaper]{article} 
\usepackage{aaai25}  
\usepackage{times}  
\usepackage{helvet}  
\usepackage{courier}  
\usepackage[hyphens]{url}  
\usepackage{graphicx} 
\usepackage{natbib}  
\usepackage{caption} 
\usepackage{amsmath}
\usepackage{amssymb}
\usepackage{booktabs}

\usepackage{algorithm}
\usepackage{algorithmic}

\usepackage{newfloat}
\usepackage{listings}
\DeclareCaptionStyle{ruled}{labelfont=normalfont,labelsep=colon,strut=off} 
\floatstyle{ruled}
\newfloat{listing}{tb}{lst}{}
\floatname{listing}{Listing}
\title{Small Vocabularies, Big Gains:\\Pretraining and Tokenization in Time Series Models}
\author{
   Alexis Roger\textsuperscript{\rm 1,2},
   Gwen Legate\textsuperscript{\rm 1, 3},
   Kashif Rasul\textsuperscript{\rm 5}, 
   Yuriy Nevmyvaka\textsuperscript{\rm 5}, 
   Irina Rish\textsuperscript{\rm 1,4}
}
\affiliations{
    \textsuperscript{\rm 1}Mila - Quebec AI Institute, 
    \textsuperscript{\rm 2}McGill University,
    \textsuperscript{\rm 3}Concordia University,
    \textsuperscript{\rm 4}University of Montreal,
    \textsuperscript{\rm 5}Morgan Stanley

}

\usepackage{bibentry}

\begin{document}

\maketitle

\begin{abstract}
Tokenization and transfer learning are two critical components in building state of the art time series foundation models for forecasting. In this work, we systematically study the effect of tokenizer design, specifically scaling and quantization strategies, on model performance, alongside the impact of pretraining versus random initialization. We show that tokenizer configuration primarily governs the representational capacity and stability of the model, while transfer learning influences optimization efficiency and alignment. Using a combination of empirical training experiments and theoretical analyses, we demonstrate that pretrained models consistently leverage well-designed tokenizers more effectively, particularly at smaller vocabulary sizes. Conversely, misaligned tokenization can diminish or even invert the benefits of pretraining. These findings highlight the importance of careful tokenization in time series modeling and suggest that combining small, efficient vocabularies with pretrained weights is especially advantageous in multi-modal forecasting settings, where the overall vocabulary must be shared across modalities. Our results provide concrete guidance for designing tokenizers and leveraging transfer learning in discrete representation learning for continuous signals.

All code, configurations and models will be released along side the paper at publication to promote future research.
\end{abstract}

%

\section{Introduction}
Time series forecasting endeavours to predict future behaviour, based on past observations, it can be univariate or multivariate. Applications are broad, they include tasks like forecasting, classification and anomaly detection and span industries from finance to healthcare \citep{ts_review}. Time series forecasting has traditionally relied on statistical models such as AutoRegressive Integrated Moving Average (ARIMA)\citep{box2015time}, Generalized AutoRegressive Conditional Heteroskedasticity (GARCH)\citep{bollerslev1986garch} and Error, Trend, Seasonal (ETS)\citep{hyndman2008forecasting}, however; these models can fail to adequately capture the long range dependencies, non-linear dynamics and domain shifts commonly present in time series data \citep{zhu2025fincast, ts_review, chronos}. With the rise in popularity of deep learning, Recurrent Neural Networks (RNNs)\citep{medsker1999recurrent}, in particular Long Short-Term Memory (LSTM)\citep{gers2000learning} models have become important time series forecasting tools. Unfortunately, despite impressive advances in the field of deep learning, these models have difficulty capturing long range dependencies and suffer from vanishing gradients \citep{hou2021stock}. \\

\noindent The standard approach to time series tokenization scales each time series by its mean followed by discretization the into uniformly spaced bins\citet{chronos}. This method wastes capacity as most bins are underutilized, while data clusters densely in narrow regions. This clustering provides benefits in the early stages of training by accelerating the learning process since many bins can be completely ignored, however; this short term advantage is quickly outweighed by the low quality of the representations. In this work we ask whether alternative scaling and binning strategies can improve token space utilization. We also examine the effect of pretraining on sensitivity of the token space distribution. Our hypothesis: is that (i) the smoother and the closer the data distribution is to normal, the better the model will perform; (ii) pretrained initialization reduces reliance on optimal tokenization. Our contributions are as follows:

\begin{itemize}
\item propose a new tokenization methods
\item show theoretical bounds of these tokenization methods
\item examines the effect of different parameters, such as the uniformity of the distribution or vocabulary size on the overall performance of the tokenizer
\item examines the effect of starting from a pre-trained initialization on the relative importance of tokenization
\end{itemize}

\section{Background and Related Work}
\subsection{Tokenization}
\noindent Tokenization is most commonly associated with natural language processing (NLP); its need arises from the inability of language models to process raw text, requiring instead and numerical input for further processing. Tokenization is acknowledged as a critical component in the development of language models, and particularly LLMs. The effectiveness of a tokenizer has a direct relationship to the ability of a model to ability to understand input and generate predictions and it therefore has a crucial role in overall model performance \citep{huangover, gastaldifoundations}. 

\subsection{Tokenization for Time Series}
\noindent Time series tokenization is the domain of time series forecasting is equally significant to language tokenization in the language modeling domain. The standard approach scales each time series by its mean and discretizes it into uniformly spaced bins\citep{chronos}. In this approach, bins are underutilized, to the detriment of training outcomes. Byte pair encoding (BPE)\citep{sennrich2016neural} is a widely adopted tokenization method in the NLP domain. It utilizes a discrete vocabulary of recurring patterns to represent language inputs. In their work, \citet{gotz2025byte} investigate byte pair encoding for time series embeddings. WaveToken\citep{masseranoenhancing}, is an innovative wavelet-based tokenizer that operates in the time-localized frequency space. The use of wavelets provides a compact representation and empirical results are strong. However, we know from studies on LLM tokenization that larger vocabularies result in better performance for the same training cost \citep{huangover} and this casts doubt on the ability of WaveToken to scale up to meet the challenge of future large time series foundation models.

\section{Theoretical Foundation}
We consider a time series $X$ governed by two main parameters: $C$ the context length and $H$ the forecasting horizon. Therefore, we have $X_{1:C}$ the historical data we use to predict $X_{C+1:H}$ Formally, we have:
$$X_{1:C} = [x_1, ~\cdots, ~x_C]  ~\text{where}~ x_i\in\mathbb{R}$$
$$X_{C+1:H} = [x_{C+1}, ~\cdots, ~x_H]  ~\text{where}~x_i\in\mathbb{R}$$
As these time series have continuous values but current LLMs have finite vocabulary, we need to discretize the real-valued observation $x_i\in\mathbb{R}$ into a finite set of tokens. This is done in two steps: scaling and quantization. It is on these steps that we want to conduct our study. 

\subsection{Scaling}
Time series from different sources can vary widely in scale, even within the same dataset, which can hinder model training. To address this, each series is independently normalized. The goal is to map the raw values into a numerically stable range suitable for quantization. This is achieved via an affine transformation of the form $\tilde{x} = ax+b$, where $a$ and $b$ are chosen according to the desired normalization scheme.

In this work, the different scaling parameters we wish to test are:

$$\text{Mean~normalization:} ~ a = \frac{1}{\frac{1}{C}\sum_{i=1}^{C}|x_i|},~ b=0$$
$$\text{Min–max~normalization:} ~ a = \frac{1}{x_{\max} - x_{\min}},~ b = -ax_{\min}$$
$$\text{Normal~normalization:} ~ a = \frac{1}{\sigma_x},~ b = -\frac{\mu_x}{\sigma_x}$$
With $\mu_x = \frac{1}{C}\sum_{i=1}^{C} x_i$ the mean of the series, and $\sigma_x = \sqrt{\frac{1}{C}\sum_{i=1}^{C} (x_i - \mu_x)^2}$ its standard deviation.

\subsection{Quantization}
After scaling, the normalized sequence 
$\tilde{x}_{1:C+H} = [\tilde{x}_1, \ldots, \tilde{x}_{C+H}]$
remains continuous and must be discretized into tokens. 
To perform this transformation, we define a set of $B$ bins with centers $c_i$ and boundaries $[b_i, b_{i+1}]$. We define the quantization function as flows:
$$
q(x) =
\begin{cases}
1, & \text{if } x < b_1, \\
2, & \text{if } b_1 \le x < b_2, \\
\vdots & \\
B, & \text{if } x \ge b_{B-1},
\end{cases}
$$

while the corresponding dequantization function $d : \{1, ~\cdots,~ B\} \rightarrow \mathbb{R}$ 
maps a discrete index back to its bin center.

\medskip

In this work, we compare three strategies for determining the bin spacing:
\begin{enumerate}
    \item \textbf{Uniform binning} — Bin centers are evenly spaced across the range $[c_1, c_B]$, and edges are positioned midway between consecutive centers:
    \[
    b_i = \frac{c_i + c_{i+1}}{2}, \quad i \in \{1, \ldots, B-1\}.
    \]
    This approach treats all regions of the value range equally and is robust to distributional shifts across datasets.

    \item \textbf{Normal binning} — Bin centers are placed according to the Cumulative Distribution Function (CDF) of a standard normal distribution. This results in finer resolution near the mean and coarser resolution in the tails, reflecting the structure of approximately Gaussian data.

    \item \textbf{Exponential-decay binning} — Bin spacing follows an exponential CDF, leading to denser bins near zero and increasingly wider bins for larger magnitudes. This configuration emphasizes small fluctuations while still covering a wide dynamic range, which can be beneficial for heavy-tailed or skewed time series.
\end{enumerate}

\begin{figure}
    \centering
    \includegraphics[width=0.9\linewidth]{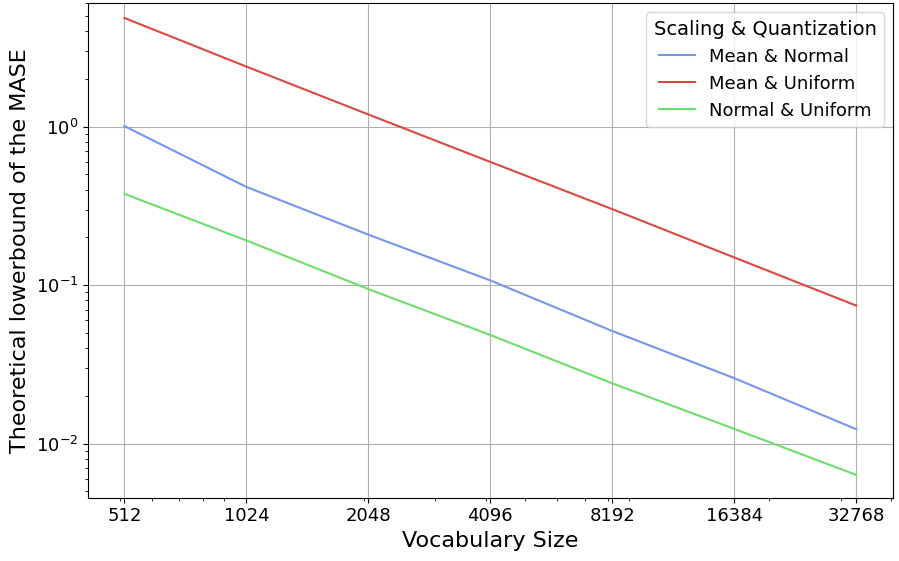}
    \caption{Graph showing how the theoretical lower-bound in performance of the scaling and quantization function combinations follows a power law with vocabulary size.}
    \label{fig:theoretical}
\end{figure}

\subsection{Finding the best combinations}
Using a perfect predictor, where the model outputs the correct bin every single time, we're able to calculate the error of each scaling and quantization configuration. This also allows us to iterate and tune the quantization hyper-parameters very rapidly and efficiently. The width of our quantization method, so the difference between $c_1$ and $c_B$, is the most important parameter, as it is a tradeoff between having more resolutions around the mean of the distribution with larger errors on the tails, or having more resolution at the tails and slightly more error near the mean.

After optimizing the parameters for all 9 scaling and quantization combinations, we evaluated them across a range of token counts. We found two that stood out by their performance and robustness: mean scaling with normal binning and normal scaling with uniform binning. We also keep mean scaling with uniform binning as a baseline as it is what was used by \cite{chronos}. The comparisons between these methods can be seen in Figure \ref{fig:theoretical}.

Figure \ref{fig:theoretical} clearly shows the power law existing between the vocabulary size and theoretical boundary. It is also interesting to see that although these are different tokenizers, they have the same slope and hence the same scaling behavior. This shows the importance of starting with the best tokenizer possible, as scaling the vocabulary up or down will maintain this advantage.

\section{Training Results}
To empirically validate the theoretical findings presented above, we trained a suite of models across varying vocabulary sizes and tokenization schemes. Specifically, we instantiated three scaling–quantization combinations—\textbf{mean \& uniform}, \textbf{mean \& normal}, and \textbf{normal \& uniform}, each with vocabularies of 512, 1024, and 4096 tokens.

All models are based on the Qwen 3 (600M) architecture \cite{qwen3technicalreport} and trained on the GiftEval pretraining dataset \cite{aksu2024giftevalbenchmarkgeneraltime}, following the standard temporal masking objective. Hyperparameters, including learning rate, batch size, and scheduler parameters, were optimized individually for each configuration to ensure a fair comparison. To isolate the effect of tokenization, we report results both from models initialized from the pretrained Qwen LLM weights and from random initialization.

\subsection{Evaluation Metric}
We adopt the Mean Absolute Scaled Error (MASE) as our primary evaluation metric, consistent with the GiftEval leader-board \cite{aksu2024giftevalbenchmarkgeneraltime} and established forecasting literature. Formally:

$$
MASE = \frac{mean(|Y-\hat{Y}|)}{seasonal\_error}
$$

where $Y$ is our prediction, $\hat{Y}$ is the expected value and \textit{seasonal\_error} denotes the mean absolute difference of the in-sample one-step naive forecast. MASE offers scale-independent interpretability and penalizes both over- and under-forecasting symmetrically.

\subsection{Empirical Trends}

\begin{figure}[b]
    \centering
    \includegraphics[width=0.95\linewidth]{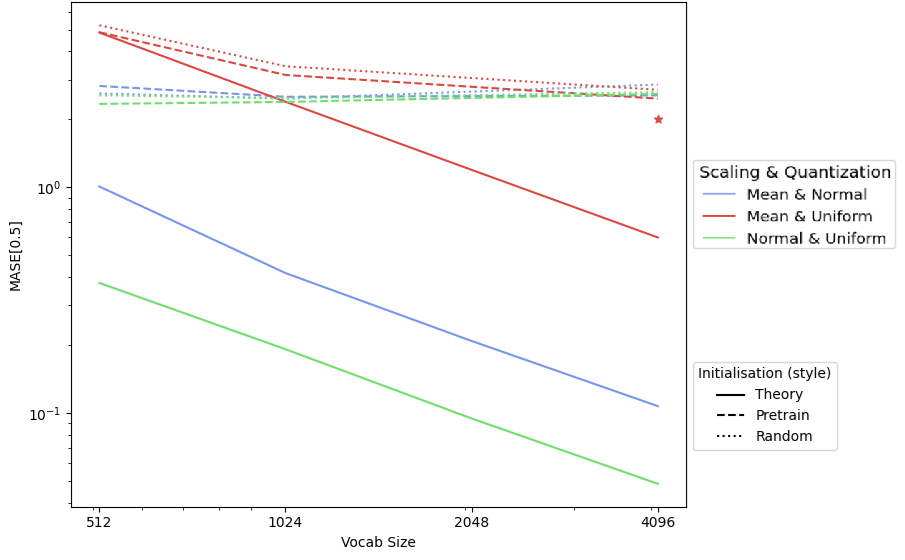}
    \caption{Plot showing the MASE score of the models at the end of training as a function of their vocab size. As performance reference, the Chronos T5 Large model, the closest to this training regime, is denoted by the red star.}
    \label{fig:mase_results}
\end{figure}

Figure~\ref{fig:mase_results} summarizes the final MASE scores across configurations. The results closely mirror the theoretical expectations:

For larger vocabularies (4096 tokens), the three tokenization schemes converge to nearly identical performance, indicating that the discretization bottleneck diminishes as representational capacity increases.

As the vocabulary size decreases, differences between schemes become increasingly pronounced. At 1024 tokens, both mean\&normal and normal\&uniform configurations surpass the theoretical lower bound for the baseline mean\&uniform tokenizer.

At the smallest vocabulary (512 tokens), the mean\&uniform configuration matches its higher theoretical limit, while mean\&normal exhibits a slight degradation and normal\&uniform maintains a measurable performance advantage.

These results confirm that the theoretical tradeoffs between scaling and quantization extend directly to practical model training. Notably, the normal\&uniform configuration consistently yields robust generalization across vocab sizes, aligning with its theoretically balanced resolution between the distribution center and tails.

While the absolute performance remains largely governed by the capacity and inductive biases of the underlying model, the tokenizer configuration critically determines how effectively the model can utilize that capacity. In particular, at smaller vocabularies—where discretization error dominates—the choice of scaling and quantization can introduce or mitigate significant performance bottlenecks.

\section{Correlation with Token-space Utilization}
To better understand the link between tokenizer quality and downstream performance, we analyzed the relationship between forecasting accuracy and the degree of token space utilization. For each tokenizer configuration, we computed Cramér’s V \cite{cramer1946mathematical}, a measure of association derived from the chi-squared statistic, to quantify how evenly the available token vocabulary is used across the dataset.

\subsection{Cramér’s V}
Cramér’s V measures the strength of association between two categorical variables and ranges from 0 (no association) to 1 (perfect association). In our context, it captures how uniformly the quantized tokens are distributed:
$$
V = \sqrt{\frac{\chi^2}{n(k - 1)}}
$$

where $\chi^2$ is the chi-squared statistic computed from the contingency table of token frequencies, $n$ is the number of samples, and $k$ is the number of token categories. A higher $V$ value indicates more balanced usage of the available tokens, suggesting better coverage of the signal’s dynamic range by the tokenizer.

\begin{table}[ht]
\centering
\small
\caption{Correlation coefficients between token space utilization (Cramér’s V) and forecasting performance (MASE). Negative correlations indicate that better token utilization corresponds to improved performance.}
\begin{tabular}{lcccccc}
\toprule
\textbf{Initialization} & \textbf{Vocab} & \textbf{Spearman} & \textbf{p} \\
\midrule
Pretrain & 512  & 1.0 & 0.0 \\
Pretrain & 1024 & 1.0 & 0.0 \\
Pretrain & 4096 & -1.0 & 0.0 \\
Random & 512  & 1.0 & 0.0 \\
Random & 1024 & 0.5 & 0.667 \\
Random & 4096 & 0.5 & 0.667 \\
Theory & 512  & 1.0 & 0.0 \\
Theory & 1024 & 1.0 & 0.0 \\
Theory & 4096 & 1.0 & 0.0 \\
\bottomrule
\end{tabular}
\label{tab:correlation}
\end{table}

\subsection{Correlation Analysis}
We then correlated the MASE score with Cramér’s V across vocab sizes and initialization strategies, as shown in Table~\ref{tab:correlation}. Since both variables exhibit monotonic but not necessarily linear relationships—particularly across small sample regimes—we employ the Spearman rank correlation coefficient to assess the strength and direction of this association. Spearman’s $\rho$ is more robust to nonlinearities and rank-based fluctuations, making it more appropriate for small-scale empirical evaluations. A visual representation of the token distributions for each tokenizer and vocabulary size is shown in Appendix \ref{app:distrib}

Table~\ref{tab:correlation} reports the Spearman rank correlations between token space utilization (Cramér’s V) and forecasting performance (MASE) across initialization regimes and vocabulary sizes. The results reveal a consistent and statistically significant monotonic relationship across nearly all settings.

For both the pretrained and theoretical predictors, we observe a strong positive correlation ($\rho = 1.0$, $p < 0.05$) at smaller vocabulary sizes (512 and 1024), indicating that higher token utilization aligns with lower forecasting error. This suggests that when the tokenizer efficiently exploits its available discrete space, the model can represent temporal dynamics more precisely, yielding better predictive accuracy.

Interestingly, at the largest vocabulary (4096), the correlation inverts ($\rho = -1.0$) for models using a pretrained initialization. This reversal suggests a saturation effect: once the token space becomes sufficiently large, excessive dispersion of token usage may lead to fragmentation, reducing statistical efficiency and model generalization. In this regime, more concentrated token utilization can actually yield better performance.

For models trained from random initialization, the correlations are weaker and less consistent ($\rho=0.5$, $p > 0.6$). This indicates that without pretrained representations, the model cannot reliably leverage the tokenizer’s structural advantages, confirming that pretraining acts as an enabler for effective token utilization.

Overall, these results reinforce the hypothesis that balanced and efficient token space utilization is necessary for forecasting performance, but also highlight a nuanced interaction between vocabulary size and model initialization. At moderate vocabularies, utilization directly drives accuracy; while at very large vocabularies, representational fragmentation may counteract this benefit.

\section{The Effects of Transfer Learning}

Transfer learning plays a central role in modern language modeling, enabling models to reuse high-level representations learned from large-scale pretraining. To quantify its impact in the context of time series tokenization, we compare models trained from random initialization against those initialized from pretrained Qwen 3 (600M) weights. Table~\ref{tab:transfer_results} summarizes the final training loss and MASE scores across scaling–quantization configurations and vocabulary sizes.

\begin{table}[h]
\centering
\small
\begin{tabular}{lccccc}
\toprule
\textbf{Initialization} & \textbf{Vocab} & \textbf{Scaling} & \textbf{Quantization} & \textbf{Loss} & \textbf{MASE} \\
\midrule
Pretrain & 4096 & Mean & Normal & 4.258 & 2.554 \\
Pretrain & 4096 & Mean & Uniform & 2.653 & 2.473 \\
Pretrain & 4096 & Normal & Uniform & 5.161 & 2.594 \\
Pretrain & 1024 & Mean & Normal & 3.058 & 2.519 \\
Pretrain & 1024 & Mean & Uniform & 1.700 & 3.147 \\
Pretrain & 1024 & Normal & Uniform & 3.839 & 2.389 \\
Pretrain & 512 & Mean & Normal & 2.554 & 2.813 \\
Pretrain & 512 & Mean & Uniform & 1.315 & 4.868 \\
Pretrain & 512 & Normal & Uniform & 3.212 & 2.342 \\
\midrule
Random & 4096 & Mean & Normal & 4.348 & 2.853 \\
Random & 4096 & Mean & Uniform & 2.683 & 2.705 \\
Random & 4096 & Normal & Uniform & 5.233 & 2.639 \\
Random & 1024 & Mean & Normal & 3.083 & 2.473 \\
Random & 1024 & Mean & Uniform & 1.697 & 3.440 \\
Random & 1024 & Normal & Uniform & 3.852 & 2.481 \\
Random & 512 & Mean & Normal & 2.550 & 2.606 \\
Random & 512 & Mean & Uniform & 1.326 & 5.226 \\
Random & 512 & Normal & Uniform & 3.233 & 2.557 \\
\bottomrule
\end{tabular}
\caption{Comparison of pretrained versus randomly initialized models across scaling–quantization configurations and vocabulary sizes. Lower MASE indicates better forecasting accuracy.}
\label{tab:transfer_results}
\end{table}

\subsection{General Trends}

\begin{figure}
    \centering
    \includegraphics[width=\linewidth]{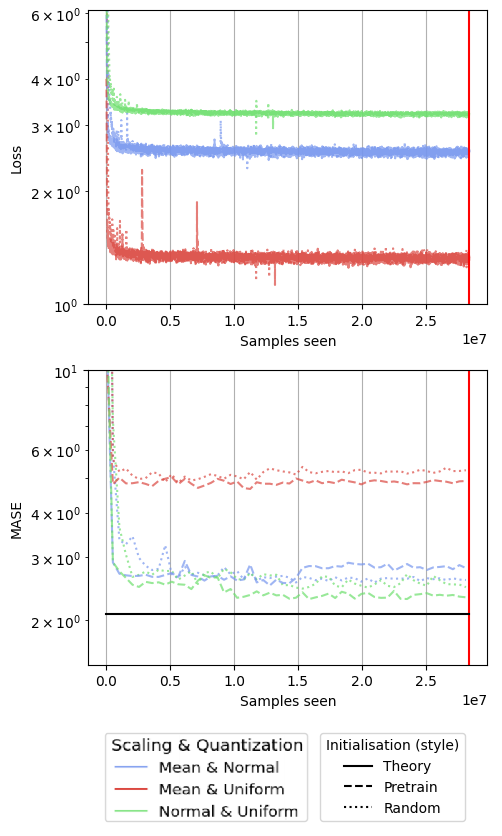}
    \caption{Loss and MASE plot across training of the different tokenization schemes for Qwen 3 600M model with a vocabulary size of 512. The vertical line to the right denotes the end of training. The bottom line indicates the performance of the Chronos T5 Large model for reference, the closest to this training regime.}
    \label{fig:training}
\end{figure}

Across all vocabulary sizes, pretrained models consistently achieve lower or comparable MASE relative to randomly initialized ones, confirming the benefit of transfer learning for time series modeling. The improvement is most pronounced for the more challenging scaling–quantization combinations and smaller vocabularies, where learning stable representations from scratch is inherently more difficult. It can also be noted that pretrained models will converge faster than randomly initialized models, even with smaller learning rates. We will focus on MASE more than loss, as loss inversely correlates with the MASE, showing its irrelevance. Figure \ref{fig:training} shows this phenomena with each tokenization.

\paragraph{Large vocabularies (4096).}
At this scale, the impact of initialization is modest but consistent. The pretrained models achieve slightly lower MASE (2.47–2.59) than their randomly initialized counterparts (2.64–2.85). This suggests that with sufficient representational capacity, pretrained knowledge primarily improves convergence speed and calibration rather than ultimate accuracy. Among the tokenizers, mean\&uniform again performs best, aligning with its strong theoretical lower bound.

\paragraph{Medium vocabularies (1024).}
Here, the benefits of pretraining become clearer. The pretrained normal\&uniform model achieves the best MASE (2.389), outperforming both the pretrained mean\&normal (2.519) and the baseline mean\&uniform (3.147). In contrast, the randomly initialized models exhibit higher variability and weaker separation between tokenizers, with the same configuration achieving 2.481 MASE but requiring longer convergence and higher loss. This indicates that pretrained weights facilitate the model’s ability to exploit nuanced tokenization structures more effectively.

\paragraph{Small vocabularies (512).}
At this smallest token budget, transfer learning provides a decisive advantage. The pretrained normal\&uniform configuration achieves the lowest MASE (2.342), whereas the randomly initialized equivalent yields 2.557. The pretrained mean\&uniform configuration performs significantly worse (4.868 MASE), matching its theoretical limitation. This contrast underscores that when discretization imposes a severe information bottleneck, pretrained representations compensate by encoding prior inductive structure that would otherwise be lost. This training run is shown in Figure \ref{fig:training}.

\section{Discussion and Conclusion}

In summary, both tokenizer design and initialization strategy critically shape model performance, but in different ways. Tokenizer configuration determines the representational capacity and stability of the model, while transfer learning influences optimization efficiency and alignment. Pretraining helps stabilize learning when vocabulary and data distributions are consistent with the pretraining regime, but provides little benefit,  or even harm, when the tokenizer diverges significantly. On small vocabularies, pretraining amplifies the effectiveness of well-designed tokenizers. This is particularly useful as reducing the vocabulary is critical for different applications, such as for multi-modal time series models, where the complete model vocabulary must be divided amongst the different modalities. To this end, we recommend future research focuses on a tokenizer using the normal scaling and uniform quantization.

Future work should explore adaptive tokenization strategies that co-evolve with pretrained embeddings, as well as cross-modal pretraining schemes explicitly robust to vocabulary shifts. Together, these directions may bridge the current gap between efficient token design and robust transfer learning for structured, discrete representations.

\subsection{Limitations}
While the experimental trends are consistent, several limitations warrant caution in interpretation:
\begin{enumerate}
    \item Synthetic task setting — The evaluation setup, though controlled, may not reflect the complexity of real temporal data distributions.
    \item Single pretraining source — All pretrained weights originated from the same corpus and architecture, leaving open questions about the generality of these observations across architectures.
    \item Unreachable theoretical baselines — The “theory” tokenizers yield extremely low MASE values but are not directly comparable since they bypass learning dynamics.
    \item It is conceivable that the ideal tokenizer could be task dependent. And this is to be kept in mind for real-world applications of these results. That being said, this methodology can be followed and replicated on a different evaluation dataset to find the optimal tokenizer.
\end{enumerate}

\section{Acknowledgments}
The authors acknowledge the AMD University Program AI \& HPC Fund for providing the high-performance computing resources used in this research. Access to these computational resources was instrumental in enabling the large-scale experiments and analyses reported in this paper.

\bigskip

\bibliography{aaai25}

\newpage

\section{Appendix}
\section{Plots showing the token distributions for the different tokenization schemes}
\label{app:distrib}
\begin{figure*}
    \centering
    \includegraphics[width=\linewidth]{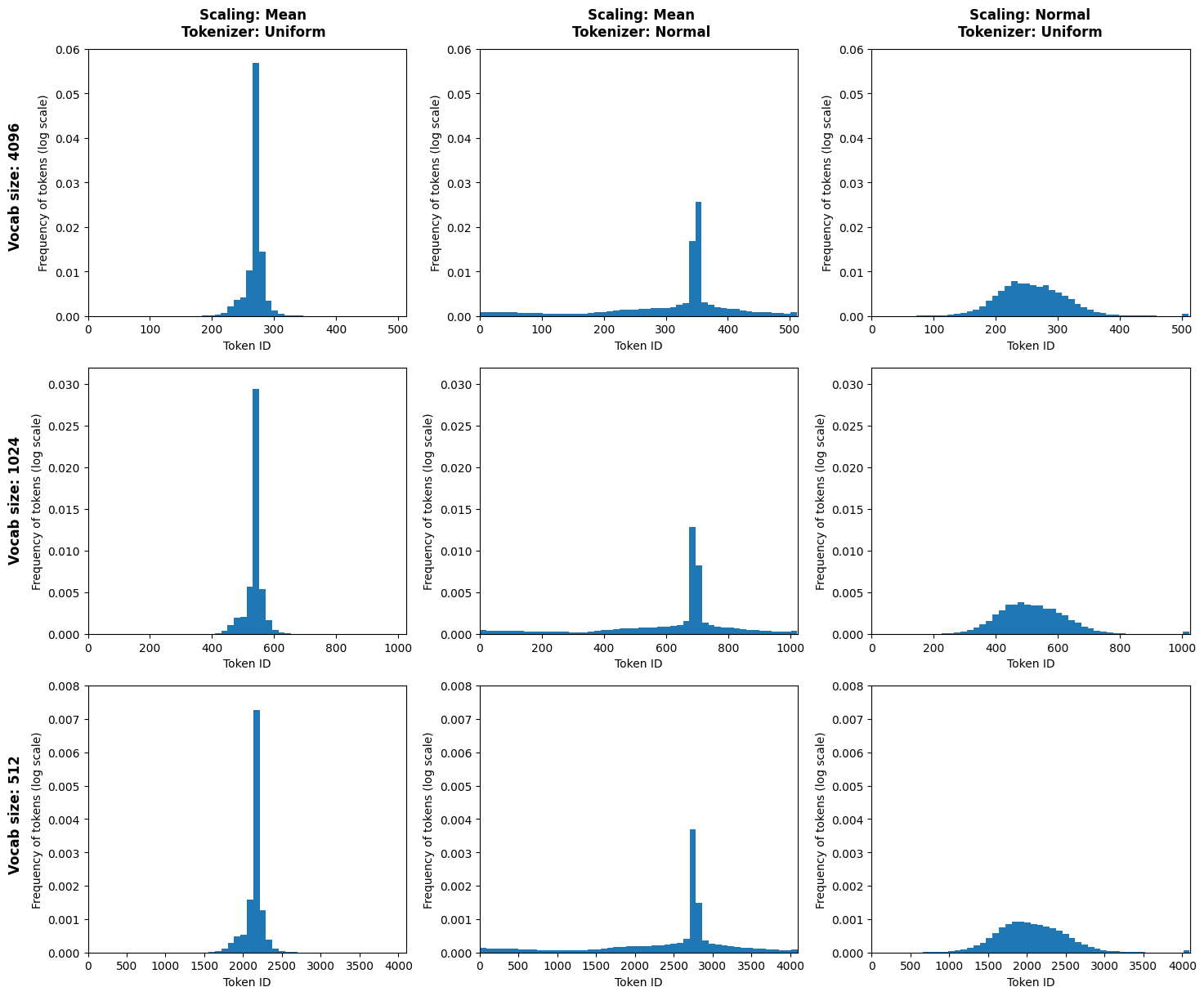}
    \caption{Histograms showing the token space utilization of the different tokenization strategies accross multiple vocabularies.}
    \label{fig:distrib}
\end{figure*}

\end{document}